\def\year{2015}
\documentclass[letterpaper]{article}
\usepackage{aaai}
\usepackage{times}
\usepackage{helvet}
\usepackage{courier}
\usepackage{times}
\usepackage{xcolor}
\usepackage{soul}
\usepackage[utf8]{inputenc}
\usepackage[small]{caption}
\usepackage{subcaption}
\usepackage[protrusion=true,expansion=true]{microtype} 

\newenvironment{talign}
 {\align}
 {\endalign}

\title{Knowledge Compilation with Continuous Random Variables\\ and its Application in Hybrid Probabilistic Logic Programming}

\author{Pedro Zuidberg Dos Martires \and Anton Dries \and Luc De Raedt  \\
KU Leuven, Belgium}

\usepackage{amsmath}
\usepackage{amsfonts}
\usepackage{amsthm}
\usepackage{txfonts}
\usepackage{graphicx, wrapfig}
\usepackage{dsfont}
\usepackage{url} 
\usepackage{amssymb}

\theoremstyle{definition}

\theoremstyle{definition}
\newtheorem{defin}{Definition}

\theoremstyle{definition}
\newtheorem{theo}{Theorem}

\theoremstyle{definition}
\newtheorem{lemma}{Lemma}

\newcommand{\lucskip}[1]{{\color{blue}(skipped text)}}

\newcommand{\pedroskip}[1]{{\color{green}(skipped text)}}

\begin{document}

\maketitle

\begin{abstract}
In probabilistic reasoning, the traditionally discrete domain has been elevated to the hybrid domain encompassing additionally continuous random variables.  Inference in the hybrid domain, however, usually necessitates to condone trade-offs on either the inference on discrete or continuous random variables.  We introduce a novel approach based on weighted model integration and algebraic model counting that circumvents these trade-offs.  We then show how it supports knowledge compilation and exact probabilistic inference. Moreover, we introduce the hybrid probabilistic logic programming language  HAL-ProbLog, an extension of ProbLog, to which we apply our inference approach.
\end{abstract}

\section{Introduction}

\noindent 
One of the state-of-the art methods for probabilistic inference in graphical models and probabilistic programming reduces probabilistic inference to weighted model counting (WMC) \cite{CHAVIRA2008772}, and then employs WMC solvers based on knowledge compilation (KC) techniques \cite{Darwiche:2002:KCM:1622810.1622817}.  Because weighted model counting applies only
to discrete probability distributions, it has recently been extended towards weighted model integration (WMI) \cite{belle2015probabilistic,morettin2017efficient} as to support also continuous distributions. However, to the best of the authors' knowledge, knowledge compilation has not yet been used for weighted model integration. 
Indeed, current approaches to weighted model integration  essentially use satisfiability modulo theory (SMT) solvers, usually restricted to linear arithmetic over the reals (SMT(${\cal LRA}$)). On the other hand, knowledge compilation has proven to be very effective, especially when many probabilistic queries need to be answered as the theory needs to be compiled only once.

The key contribution of this paper is that we show how standard KC techniques can be applied to solve weighted model integration problems, and that we incorporate such techniques in hybrid probabilistic logic programming languages. 
This is realized by  casting weighted model integration within the framework of algebraic model counting (AMC) \cite{kimmig2017algebraic}. AMC generalizes the standard weighted model counting problem to work with arbitrary semirings.   More specifically, we make the following contributions:
\begin{enumerate}
\item
We introduce the probability density semiring. 
\item 
We show how this allows to cast WMI within AMC and thereby to use the general body of literature on knowledge compilation.
\item
We introduce \emph{Symbo}, a solver for WMI that realizes knowledge compilation and exact symbolic inference.
\end{enumerate}
Symbo exploits results by \cite{gehr2016psi} to simplify the algebraic expressions.

Algebraic model counting has also been incorporated in  logic programming languages such as aProbLog \cite{kimmig2011algebraic}, which is an extension of the probabilistic programming language ProbLog \cite{fierens2015inference} towards semirings and allows to state our next contribution.
\begin{enumerate}
\setcounter{enumi}{3}
\item
We use the probability density semiring $S$ within aProbLog 
to obtain a hybrid probabilistic logic programming language and implementation, that we dub \textit{\textbf{h}ybrid \textbf{al}gebraic \textbf{ProbLog}} or HAL-ProbLog
\end{enumerate}
HAL-ProbLog is related to the Distributional Clauses framework of \cite{Gutmann2011,gutmann2011magic,nitti2016probabilistic}, differences and similarities are further discussed in subsection \ref{sec:hal_problog_syntax_semantics}.

\section{Preliminaries}\label{sec:prelim}
\subsection{Weighted Model Integration}
While the well-known SAT problem is the problem of deciding whether there is a satisfying assignment to a logical formula or not, an SMT problem generalizes SAT and allows in addition to use expressions formulated in a background theory. Consider, for example, the following SMT theory $\mathtt{broken}$:
\begin{align}\label{eqn:broken}
\mathtt{broken} \leftrightarrow ( \mathtt{no\_cool} \land (\mathtt{t}> 20) )  \lor (\mathtt{t}> 30) 
\end{align}
where $\mathtt{no\_cool}$ is a Boolean variable, while $\mathtt{t}$ is a real-valued variable. SMT then answer the question whether or not there is a satisfying assignment to the formula for the variables $\mathtt{no\_cool}$ and $\mathtt{t}$. In this paper we consider non-linear real arithmetic SMT formulas.

\begin{defin}\label{def:smtnra} (SMT($\mathcal{NRA}$) (non-linear real arithmetics)) Let $\mathbb{R}$ denote the set of real values and $\mathbb{B}=\{0,1\}$ the set of Boolean values. We then define SMT($\mathcal{NRA}$) theories as combinations by means of the standard Boolean operators $\{\neg, \land, \lor, \rightarrow, \leftrightarrow \}$ of atomic propositions $a_i \in \mathbb{B}$ and of $\mathcal{NRA}$ atomic formulas in the form $\sum_i c_i \cdot x_i^{p_i}\lesseqgtr c$. The $x_i$ are variables in $\mathbb{R}$ and $c_i,c, p_i \in \mathbb{Q}$. Allowing only for $p_i=1$ restricts SMT($\mathcal{NRA}$) theories to  SMT($\mathcal{LRA}$ (linear real arithmetics) theories. \qed
\end{defin}

Consider again the theory $\mathtt{broken}$ (cf. Eq. \ref{eqn:broken}). Assume that $\mathtt{t}$ is distributed  according to: $\mathtt{t} \sim \mathcal{N}_\mathtt{t}(20,5)$ and that the probability for $\mathtt{no\_cool}$ being true is $0.01$. Determining now the probability of the formula being true extends the SMT problem to weighted model integration. We introduce WMI following \cite{morettin2017efficient}.
\begin{defin}(Weighted model integration (WMI))\label{def:wmi} Given is a set $B$ of $M$ Boolean variables,  a set $X$ of $N$ real variables, a weight function $w(X,B): \mathbb{B}^M \times \mathbb{R}^N \rightarrow \mathbb{R}^+$, and an SMT formula $\phi(X,B):  \mathbb{B}^M \times \mathbb{R}^N \rightarrow \mathbb{B}$, the {\bf weighted model integral (WMI)} is
\begin{align}\label{eqn:wmi}
 \textstyle WMI(\phi,w \mid X, B) = \sum_{{\bf b}\in {\cal I}_B(true)} \int_{{\bf x} \in {\cal I}_{X}(\phi(X,{\bf b}))} w({\bf x},{\bf b}) d{\bf x}
\end{align}
where we use the notation ${\cal I}_V( \phi(V) )$ to denote the set of 
assignments to the variables in $V$ that satisfy $\phi(V)$.\qed
\end{defin}

Hence, when computing the weighted model integral (Eq. \ref{eqn:wmi}), we first integrate over all $X$ in a formula $\phi(X, {\bf b})$ for each possible assignment ${\bf b}$ to the Boolean variables $B$ holds and then sum up the values of the integrals.
The weight function is used to map a set of variable assignments to their weight. The weight function usually {\em factorizes} as the product of the weights over the different variables, i.e.,
$w({\bf x},{\bf b}) = \prod_{x_i \in {\bf x}} w(x_i) \prod_{b_j \in {\bf b} } w(b_j)$.

With the definition of WMI at hand, we can also produce the weighted model integral by integrating over the continuous random variables, while using the algebraic constraints as boundary conditions and weighting the integrals with the probability of the Boolean variables: $WMI(\mathtt{broken})\text{=}0.01\int_{20<\mathtt{t}\leq 30} {\cal N}_\mathtt{t}(20,5)d\mathtt{t}\text{+}\int_{\mathtt{t}>30} {\cal N}_\mathtt{t}(20,5)d\mathtt{t}$.

\subsection{Algebraic Model Counting}
\begin{defin} (Weighted model counting (WMC)). WMC is  the special case of weighted model integration where the set of real variables is empty: $X=\emptyset$.\qed
\end{defin}
WMC is traditionally used for probabilistic inference in Bayesian networks \cite{CHAVIRA2008772} and probabilistic programming \cite{fierens2015inference} with a factorized weight function:
$\textstyle WMC(\phi, w |B) = \sum_{{\bf b} \in {\cal I}_B(\phi(B))}  ~~ \prod_{b_i \in {\bf b}} w(b_i)$.
Algebraic model counting \cite{kimmig2017algebraic} generalizes WMC to commutative semirings.
More formally, 
\begin{defin}\label{def:comm_semiring}
 A {\bf  commutative semiring} is an algebraic structure $(\mathcal{A},\oplus,\otimes,e^{\oplus},e^\otimes)$ such that (1) addition $\oplus$ and multiplication $\otimes$ are binary operations $\mathcal{A}\times \mathcal{A}\rightarrow \mathcal{A}$; (2)  addition $\oplus$ and multiplication are  associative and commutative binary operations over the set $\mathcal{A}$; (3) $\otimes$ distributes over $\oplus$; (4) $e^\oplus \in \mathcal{A}$ is the neutral element of $\oplus$; (5) $e^\otimes \in \mathcal{A}$ is the neutral element of $\otimes$; and (6) $e^\oplus$ is an annihilator for $\otimes$.\qed
 \end{defin}
\begin{defin}(Algebraic model counting (AMC)) Given:
\begin{itemize}
\item a propositional logic theory $\phi$ over a set of variables $B$
\item a commutative semiring $(\mathcal{A},\oplus,\otimes,e^{\oplus},e^\otimes)$
\item a labeling function $\alpha:\mathcal{L}\rightarrow \mathcal{A}$, mapping literals $\mathcal{L}$ from the variables in $B$ to values from the semiring set $\mathcal{A}$
\end{itemize} 
The algebraic model count of a theory $\phi$ is then defined as:
\begin{align} \label{eqn:labelamc}
\textstyle AMC(\phi,\alpha | B) = \bigoplus_{{\bf b} \in \mathcal{I}_B(\phi(B))} \bigotimes_{b_i \in {\bf b}} \alpha (b_i)& &\qed \nonumber
\end{align}
\end{defin}
We use $\alpha$ instead of $w$ and the term label rather than weight to reflect that the elements of the semiring cannot always be interpreted as weights.

\cite{kimmig2017algebraic} show also under which conditions the an algebraic model count is a valid computation.
\begin{defin}\label{lem:nsp}(Neutral-sum property) A semiring addition and labeling function pair $(\oplus,\alpha)$ is neutral \emph{iff.} $\forall b \in B:  \alpha(b)\oplus \alpha(\neg b)=e^\otimes$.\qed
\end{defin}
\begin{theo}\label{theo:amc_ddnnf}(AMC on d-DNNF) Evaluating a d-DNNF representation of the propositional theory $\phi$, using Algorithm 1 in \cite{kimmig2017algebraic}, for a semiring and labeling function with neutral tuple $(\oplus, \alpha)$ is a correct computation of the algebraic model count, cf. \cite{kimmig2017algebraic}. \qed
\end{theo}

\subsection{Knowledge Compilation}
Knowledge compilation \cite{Darwiche:2002:KCM:1622810.1622817} is the process of transforming a propositional logic formula  into a form that allows for polytime evaluation of the formula. Although the knowledge compilation step itself is computationally hard, the overall procedure yields a net benefit when a logical circuit has to be evaluated multiple times, possibly with different labels/weights for the literals.

A popular language to compile propositional formulas into are Sentential Decisions Diagrams (SDDs) \cite{Choi2013}. SDDs are s a subset of d-DNNF formulas. We use SDDs to implement our solver, Symbo.

Note that, as SDDs are subset of d-DNNF, Theorem \ref{theo:amc_ddnnf} holds also for the them.

\section{The probability density semiring and WMI}
 
Now we have all the ingredients to define the {\em probability density semiring}, which is needed to cast WMI as AMC.

The key difference between WMI and AMC is that in a WMI task there is first a sum, then an integral and then typically a product, while in AMC there is no integral. 
This intuitively implies that, if we want to 
cast WMI using AMC, we will have to perform the integration last: WMI = $\int$ AMC.
This can only be realized if we keep track of the two elements needed in the integral 1) the formula $\phi$ defining the values over which to integrate and 2) the weight function $w$ defining the densities according to which the variables in $\phi$ are distributed. 
So, the set of elements of semiring that we need to define will
consist of tuples, where the first element will denote an algebraic expression and the second the weight function.
\begin{defin}\label{def:label_alpha} (Labeling function $\alpha$)
If the literal $l$ represents either a Boolean variable $b$ or its negation $\neg b$ then the label 
\begin{align}
\alpha(b) \coloneqq \left( P(b) ,  \emptyset \right) &&\alpha (\neg b)  \coloneqq \left( 1-P(b) ,  \emptyset \right)
\end{align}
where $P(b)$ denotes the probability of the Boolean variable. Otherwise if the literal $l$ corresponds to an algebraic constraint within  SMT(${\cal NRA}$), depending on the set of real-valued continuous random variables $\{ \bf x \}$, then the label of $l$ is  given by:
\begin{align}
\alpha(l) \coloneqq \left([l] , \mathcal{F}^S_{x}\right)  \label{eqn:label_halproblog} &&
\alpha (\neg l) \coloneqq   \left([\neg l] , \mathcal{F}^S_{x}\right)
\end{align}
 $\mathcal{F}^S_{x}$ denotes the set $\left \{ x_i \text{$\sim$} f_{i} \right \}$, where the $x_i$ are random variables and the $f_i$ the corresponding probability densities. The first definition in  Eq. \ref{eqn:label_halproblog} is read as {\em `l such that any $x_i$ is distributed according to the corresponding $f_i$'}. \qed
\end{defin}
The brackets around $[l]$ denote the so-called \textit{Iverson brackets}. They evaluate to $1$ if their argument $l$ evaluates to true and to $0$ otherwise - they are a generalized indicator function.

We can now define the set of elements of the semiring:
\begin{defin}(Probability density semiring $\mathcal{S}$) The elements of the semring $\mathcal{S}$ are given by the set
\begin{equation}\label{eqn:set}
\textstyle \mathcal{A}\coloneqq \left \{ \left(a,\mathcal{F}^S_{x} \right) \right \}
\end{equation}
where $a$ denotes any algebraic expression over ${\cal NRA}$, including also Iverson brackets. For instance, $a=0.01[20 < t\leq 30] + [t > 30]$. $\mathcal{F}^S_{x}$ is  shorthand for the set $\left \{ x_i \sim f_{i} \right \}$ with $x$ the set of all real-valued continuous random variables appearing in $a$. \\
The neutral elements $\oplus$ and $\otimes$ are defined as:
\begin{align}
e^\oplus  \coloneqq (0, \emptyset)  &&
e^\otimes  \coloneqq (1, \emptyset)
\label{eqn:hal_problog_neutral}
\end{align}
For the addition and multiplication we define:
\begin{align}
\left(a_1,\mathcal{F}^S_{x_1}\right) \oplus
 \left(a_2,\mathcal{F}^S_{x_2} \right)
&\coloneqq \left(a_1+a_2,\mathcal{F}^S_{x_1} \cup \mathcal{F}^S_{x_2} \right)  \label{eqn:hal_problog_sum} \\
\left(a_1,\mathcal{F}^S_{x_1} \right) \otimes
 \left(a_2,\mathcal{F}^S_{x_2} \right)
&\coloneqq \left(a_1 \times a_2,\mathcal{F}^S_{x_1} \cup \mathcal{F}^S_{x_2} \right) 
\label{eqn:hal_problog_prod}
\end{align}\qed
\end{defin}


\begin{lemma}\label{lem:iprob_comm_semiring} The structure $\mathcal{S}=({\cal A}, \oplus, \otimes, e^\oplus,e^\otimes)$ is a commutative semiring.\\
\textbf{Proof (Sketch).} To prove that the structure $\mathcal{S}$ actually constitutes a commutative semiring, we need to show that the properties in Definition \ref{def:comm_semiring} hold. 

The proof relies on the commutativity and associativity of the Iverson brackets under standard addition and multiplication, and on the commutativity and associativity of the union operator for sets. Similarly for the distributivity of the multiplication over the addition (c.f. property $3$). Lastly, properties $4$ to $6$ are trivially satisfied. We conclude that the structure ${\cal S}$ is indeed a commutative semiring. \qed
\end{lemma}

\begin{lemma}\label{lem:neutral_sum} The pair  $(\oplus, \alpha)$ is neutral.\\
\textbf{Proof.} Let $l$ be a literal with label $\alpha(l)=P(l)$ \emph{iif.} $l \in B$ and $\alpha(l)=[l]$ \emph{iff.} $l$ is an abstraction of an $\mathcal{NRA}$ formula. We then have: 
\begin{align}
\alpha(l)\oplus \alpha({\neg} l)&=
\begin{cases}
(P(l) ,\emptyset) \oplus  ( 1-P(l), \emptyset) & \text{for }l\in B\\
([l] ,\mathcal{F}) \oplus  ( [\neg l], \mathcal{F}) &  \text{else}\\
\end{cases} \nonumber \\
&= 
\begin{cases}
( 1, \emptyset)\nonumber &  \text{for }l\in B\\
( [l]+[\neg l], \mathcal{F})& \text{else}\\
\end{cases}\nonumber \\
&=e^{\otimes} &
 \nonumber
\end{align}
In the last line we used the fact that $(1,\mathcal{F})$ and $(1, \emptyset)$ are equivalent elements within the probability density semiring. \qed
\end{lemma}

\begin{lemma}\label{lem:amc_pds} (AMC on d-DNNF with $\mathcal{S}$) The algebraic model count is a valid calculation on a d-DNNF representation of a logic formula given the density semiring $\mathcal{S}$\\
{\bf Proof.} This follows immediately from Lemma \ref{lem:iprob_comm_semiring} and \ref{lem:neutral_sum}, together with Theorem \ref{theo:amc_ddnnf}. \qed
\end{lemma}
An SMT(${\cal NRA}$) theory induces an infinity of theories, one for each possible instantiation of continuous random variables. We can utilize the same compiled theory for each of the infinitely many theories. Note that the probability of each instantiation is $0$, as the support for a single instantiation is infinitesimally small. In order to retrieve actual probabilities from an SMT, we need to carry out the integration over the continuous random variables.

We show now how computing the algebraic model count in the semiring setting $\mathcal{S}$ yields the probability of a theory being satisfied.
\begin{theo}
Let $\phi$ be an SMT(${\cal NRA}$) theory, $w$ a factorized weight function over the Boolean variables $B$ and continuous variables $X$. Furthermore, assume that $AMC(\phi, w | X \cup B)$ evaluates to $(\Psi , {\cal F}^S_x)$ in the semiring ${\cal S}$, with $\Psi = \sum_{{\bf v} \in \mathcal{I}(\phi(X,B))} \prod_{v_i \in {\bf v}} a_{v_i}$, where $a_v$ is an algebraic expression depending on the random variables ${\bf v}$. Then
\begin{equation*}
 \textstyle WMI(\phi,w | X, B) = \int_{{\bf x} \in X} \Psi \prod_{f_i(x_i) \in \mathcal{F}^S_x} f_i(x_i)  d{\bf x}
\end{equation*}
$\prod_{f_i(x_i) \in \mathcal{F}^S_x} f_i(x_i)$ is the product over the probability density function of the continuous random variables appearing in $\Psi$.\\
{\bf Proof} In the first step we re-write $\Psi$ as the sum-product over the algebraic expression $a_v$. We note also that the product over the density functions is actually the weight of the continuous random variables in WMI. In the second step (\ref{proof:intamc2} to \ref{proof:intamc3}) we split up the sum and the product over the variables ${\bf v}$ into sums over the Boolean and continuous random variables - likewise for the product. Next (\ref{proof:intamc3} to \ref{proof:intamc4}) we push the product over the Boolean random variables through and note in (\ref{proof:intamc5}) that this product corresponds to the weight of the Boolean random variables in WMI.
\begin{subequations}
\begin{talign}
& \int_{{\bf x} \in X} \Psi \prod_{f_i(x_i) \in \mathcal{F}_x} f_i(x_i) d{\bf x} \label{proof:intamc1} \tag*{P1} \\
=&\int_{{\bf x} \in X} \sum_{{\bf v} \in \mathcal{I}(\phi(X,B))} \prod_{v_i \in {\bf v}} a_{v_i}  w({\bf x}) d{\bf x}  \label{proof:intamc2}\tag*{P2} \\
=&\int_{{\bf x} \in X} \sum_{{\bf b}\in \mathcal{I}_b(true)} \sum_{{\bf x} \in \mathcal{I}_X(\phi(X,{\bf b}))} \prod_{b_i\in {\bf b}} \prod_{x_i \in {\bf x}} a_{b_i} a_{x_i} w({\bf x}) d{\bf x}  \label{proof:intamc3}\tag*{P3} \\
=&\int_{{\bf x} \in X} \sum_{{\bf b}\in \mathcal{I}_b(true)} \sum_{{\bf x}\in \mathcal{I}_X(\phi(X,{\bf b}))}  \prod_{x_i \in {\bf x}} a_{x_i} \prod_{b_i\in {\bf b}}  a_{b_i} w({\bf x}) d{\bf x}  \label{proof:intamc4} \tag*{P4}\\
=&\int_{{\bf x} \in X} \sum_{{\bf b}\in \mathcal{I}_b(true)} \sum_{{\bf x} \in \mathcal{I}_X(\phi(X,b))}  \prod_{x_i \in {\bf x}} a_{x_i} w({\bf b})w({\bf x}) d{\bf x}  \label{proof:intamc5}\tag*{P5} \\
=&\sum_{{\bf b}\in \mathcal{I}_b(true)} \int_{{\bf x} \in X}   \sum_{{\bf x}\in \mathcal{I}_X(\phi(X,{\bf b}))}  \prod_{x_i \in {\bf x}} a_{x_i} w({\bf x},{\bf b}) d{\bf x}  \label{proof:intamc6}\tag*{P6} \\
=&\sum_{{\bf b}\in \mathcal{I}_b(true)} \int_{{\bf x} \in {\cal I}_{X}(\phi(X,{\bf b}))}  w({\bf x},{\bf b}) d{\bf x}  \label{proof:intamc7}\tag*{P7}
\end{talign}
\end{subequations}
In the last two lines, we exchanged the summation and integral, as Fubini's theorem holds for summations/integrals over probability distributions and densities. The integral over the so-obtained sum-product is the integral over Iverson brackets. We rewrite the indefinite integral over the Iverson brackets as the definite integral having boundary conditions corresponding to the condition present in the Iverson brackets.

The last line (\ref{proof:intamc7}) corresponds to the definition of the weighted model integral. We have, hence, shown that WMI can be cast as an AMC task.
\qed
\end{theo}

\section{Probability of SMT formulas via KC}\label{sec:kc}
We describe now {\bf Symbo}, a symbolico-logic algorithm that produces the weighted model integral of an SMT($\mathcal{NRA}$) formula $\phi$ via knowledge compilation.

In Lemma \ref{lem:amc_pds} we saw that the probability semiring $S$ can be used to calculate the algebraic model count on a d-DNNF representation of a logical formula. Recalling Theorem \ref{theo:amc_ddnnf}, we are hence also capable of obtaining the weighted model integral of the a hybrid propositional formula, given the probability distributions of the random the variables.

 At a high level, Symbo takes the following consecutive steps:
\begin{enumerate}
\item Abstraction of algebraic constraints in $\phi$ in order to obtain $\phi^{abstract}$. For instance, a constraint $(t > 20)$ would be abstracted as a Bool $b_{t>20}$.
\item Compilation of $\phi^{abstract}$ into a d-DNNF representation $\phi^{abstract}_{compiled}$.
\item Transforming the logic formula $\phi^{abstract}_{compiled}$ into an arithmetic circuit $AC_\phi$.
\item Labeling the literals in $AC_\phi$ according to the labeling function given in Definition \ref{def:label_alpha}.
\item Symbolically evaluating $AC_\phi$ according to the probability density semiring $\mathcal{S}$.
\item Symbolically multiplying the expression obtained from evaluating $AC_\phi$, which is a sum-product of weighted indicator functions (Iverson brackets), by the probability densities according to which the continuous random variables are distributed.
\item Symbolically integrating out the continuous random variables.
\end{enumerate}

Regarding more technical details of the algorithm: Symbo leverages the PSI-Solver \cite{gehr2016psi}, a novel approach for exact symbolic analysis of probabilistic programs that carries out inference through symbolic reasoning\footnote{This includes, amongst others, algebraic simplifications and guard simplifications. See \cite{gehr2016psi} for a detailed discussion.}. When evaluating a compiled hybrid theory, Symbo builds up a symbolic PSI expression for $\Psi$ (cf. Theorem \ref{theo:amc_ddnnf}). The leaf nodes of the d-DNNF representation are annotated with algebraic expressions. A leaf corresponding to a Boolean literal receives a symbolic value for the probability of being satisfied and a leaf corresponding to an abstraction of an algebraic condition is expressed as a symbolic Iverson bracket. Internal logical nodes of the compiled theory (logical and/or operations) are mapped  to the symbolic multiplication/addition of the PSI-Solver. The inference engine of the PSI-Solver tries to symbolically simplify resulting expressions as much as possible. Once the compiled circuit is evaluated, the final expression is multiplied by Symbo with the set of densities corresponding to the continuous random variables in $\Psi$. The symbolic integration is then again carried out by the PSI-Solver.

\begin{figure}
\begin{center}
\includegraphics[width=\linewidth]{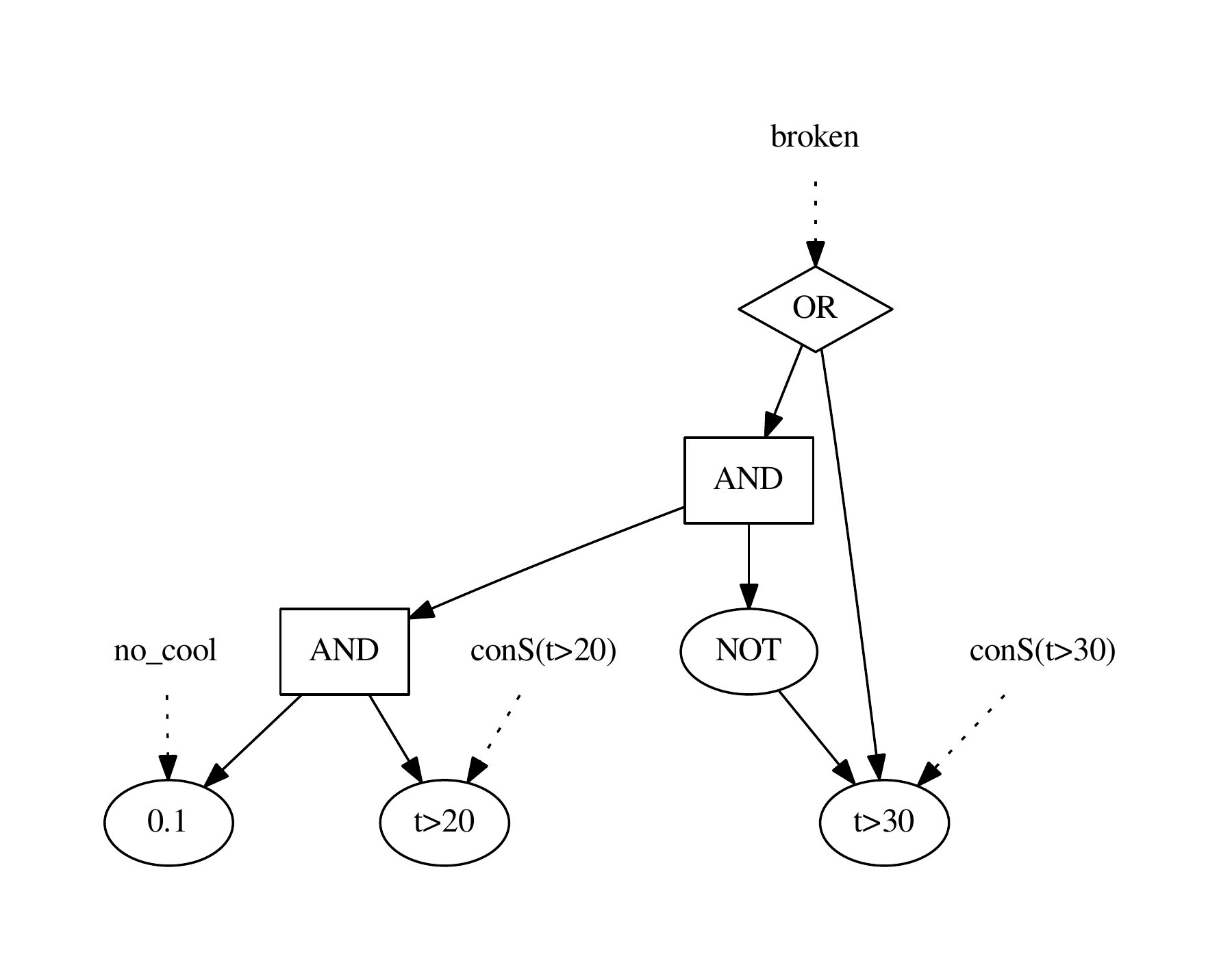}
\caption{Graphical representation of the propositional formula in Eq. \ref{eqn:broken}.}
\label{fig:sdd_broken}
\end{center}
\end{figure}

Lets look at an example. Consider our initial example in Eq. \ref{eqn:broken}. Compiling it into d-DNNF form yields:
\begin{align}
( \mathtt{no\_cool} \land (\mathtt{t}> 20) \land (\mathtt{t}\leq 30) )  \lor (\mathtt{t}> 30) 
\end{align}
which is already a propositional formula in a d-DNNF representation. Such a hybrid formula, for which Symbo has kept track of the probabilities and weights involved, can be considered to be the input to the algorithm. We can represent it as a graph where the leaves represent the literals in the formula and internal nodes logical operation, cf. Figure \ref{fig:sdd_broken}. Evaluating this theory using the semantics of the probability semiring $\mathcal{S}$ and the PSI-Solver yields the following result

$$\Psi =  0.01 [\mathtt{t}\text{$>$}20] [\mathtt{t}\text{$\leq$} 30]+ [\mathtt{t}\text{$>$}30] $$
Multiplying this expression by the  probability density function for $\mathtt{t}$ and carrying out the integral gives us the weighted model integral for the theory $\mathtt{broken}$.
\begin{align}
p(\mathtt{broken}) = & \int \left( 0.01 [\mathtt{t}\text{$>$}20] [\mathtt{t}\text{$\leq$} 30]+ [\mathtt{t}\text{$>$}30] \right) \mathcal{N}_{\mathtt{t}}(20,5) d\mathtt{t} \nonumber \\
= & 0.01\int_{20\text{$<$}\mathtt{t}\text{$\leq$}30}\mathcal{N}_{\mathtt{t}}(20,5) d\mathtt{t}\text{$+$} \int_{\mathtt{t}\text{$>$}30}\mathcal{N}_{\mathtt{t}}(20,5) d\mathtt{t} \nonumber \\
= &1 -0.01 \left( \frac{d}{dx} \right)^{-1}[e^{-x^2}]\left(-\frac{5}{2}\sqrt{8}+\frac{20}{\sqrt{8}}\right)\frac{1}{\sqrt{\pi}}\nonumber \\
&-0.9 \left( \frac{d}{dx} \right)^{-1}[e^{-x^2}] \left( -\frac{5\sqrt{8}}{2}+\frac{30}{\sqrt{8}} \right) \frac{1}{\sqrt{\pi}} \nonumber
\end{align}
In PSI, terms of the form $(d / dx)^{-1}[e^{-x^2}](a)$ denote the function $\int_{-\infty}^{a}dx e^{-x^2} $, which cannot be simplified any further.

We note that the symbolic inference engine underlying the PSI-Solver has until now only been used for imperative programing. The implementation of Symbo shows that the powerful symbolic inference engine can also be adopted for logic programming when making use of knowledge compilation.

\section{HAL-ProbLog}
Let us now define HAL-ProbLog, a hybrid probabilistic logic programming language
based on the distributional clause semantics of \cite{gutmann2011magic,nitti2016probabilistic}.
By making use of the reduction of WMC to AMC, we can implement
HAL-ProbLog as an instance of aProbLog \cite{kimmig2011algebraic}, which  itself 
extends the semantics of ProbLog \cite{fierens2015inference}, a probabilistic logic programming language. While ProbLog solves the task of computing the probability of a certain query being true, aProbLog generalizes this to a variety of other tasks by deploying 
a semiring.

\subsection{aProbLog}
\begin{defin}(aProbLog program) An algebraic ProbLog program consists of:
1) a commutative semiring $\mathcal{S}$, 2) a finite set of ground algebraic facts $F=\{ f_i \}$, 3) a finite set BK of background knowledge clauses of the form $h \leftarrow b_1, ... , b_n$ where $h$ and the $b_i$ are logical atoms, and 4) a labeling function $\alpha : \mathcal{L}(F)\rightarrow \mathcal{A}$ where $\mathcal{L}(F)$ contains all facts $f \in  F$ and their negation $\neg f$.
\end{defin}

Following \cite{kimmig2011algebraic}, we also define an aProbLog query and the label of a resulting theory as follows. 
\begin{defin}(aProbLog query) An aProbLog query $q$ is a finite set of algebraic literals and atoms from the Herbrand base ($HB$), i.e. the set of ground atoms that can be constructed from the predicates, functor and constant symbols of the program $q \subseteq \mathcal{L}(F) \cup HB(F \cup BK)$. The set of interpretations ${\cal I}(q)$ that makes the query $q$ true is defined as:
$\textstyle {\cal I}(q) = \{ I \subseteq {\cal L}(F) | \forall l \in F : l \in I \leftrightarrow \neg l \not\in I \mbox{ and }
I \cup BK \models q \}$.
\end{defin}

\begin{defin}\label{defap} (Label of aProbLog query) The label of a query $q$ is the label of $\mathcal{I}(q)$: $\textstyle \boldsymbol{AMC}(q)=\boldsymbol{AMC}(\mathcal{I}(q))=\bigoplus_{I \in \mathcal{I}(q)} \bigotimes_{l \in I} \alpha (l)$.
\end{defin}

\subsection{Syntax and semantics of HAL-ProbLog}\label{sec:hal_problog_syntax_semantics}
We now apply aProbLog to obtain HAL-Problog, which we first illustrate on a simple example
modeling the behavior of a machine under different temperature conditions. This examples is an extension of the SMT formula in Eq. \ref{eqn:broken}.
\begin{flalign}
&0.2::\mathtt{h}.&  \tag*{\%$0.2$ chance of being a hot day}  \\
&0.01::\mathtt{no\_cool}.& \tag*{\%$0.01$ chance of cooling not working} \\
&\mathtt{normal}(20,5)::\mathtt{t}\leftarrow \neg \mathtt{h}.& \tag*{\%temperature distribution} \\
&\mathtt{normal}(27,5)::\mathtt{t}\leftarrow \mathtt{h}.& \nonumber \\
&\mathtt{broken}\leftarrow \mathtt{valS}(\mathtt{t},T), \mathtt{conS}(T>30).& \nonumber \\
&\mathtt{broken}\leftarrow \mathtt{no\_cool}, \mathtt{valS}(\mathtt{t},T), \mathtt{conS}(T>20). &  \label{exmp:halproblog} 
\end{flalign}

Looking at the program in Eq. \ref{exmp:halproblog}, we observe two differences in comparison to orthodox ProbLog syntax. Firstly, we can describe not only Boolean random variables but also continuous random variables, and specify how the random variables are distributed. 
This is realized by statements of the form  $D::t\leftarrow b_1, ... , b_n$,
which denotes that $t\theta$ is a continuous random variable distributed according to $D\theta$ whenever $b_1\theta, ... , b_n\theta$ are true
for a substitution $\theta$ that grounds the rule. We will use the shorthand $t\theta |b_1\theta, ... , b_n\theta  \sim D\theta $ (we read this {\em $t$ given  $b_1$ and ... and $b_n$'}). In our example we have $\mathtt{t|not( w)} \sim \mathcal{N}(20,5)$. The temperature random variable $\mathtt{t}$  is distributed according to a specific normal distribution given it being a hot day or not.

A second difference to ordinary ProbLog lies in allowing HAL-Problog to also encompasses conditional statements - allowing to define binary random variables that depend on continuous ones. Therefore, we utilize the two built-in predicates $\mathtt{valS}/ 2$ and $\mathtt{conS}/ 1$. $\mathtt{valS}/ 2$ takes as first argument a variable and the second argument unifies with a symbol representing the value of the variable. The $\mathtt{conS}/ 1$ predicate denotes an Iverson bracket involving symbolic values. Note that HAL-ProbLog allows to deploy conditions such as the following: $(\mathtt{t>r}+10)$ and $(\mathtt{t}^3>e^\mathtt{r})$. Whether programs involving such expression can be solved or not relies on the solver. Using Symbo as a solver only programs reducible to $\mathcal{NRA}$ formulas are guaranteed to be solvable.

In order to obtain meaningful HAL-ProbLog programs, each possible world allows for only one possible definition of one and the same continuous random variable, this construct is similar to that of the Distributional Clauses \cite{nitti2016probabilistic,Gutmann2011}. This effectively means that, exactly as Distributional Clauses, we only allow for mixtures of continuous random variables and is guaranteed by requiring that rules with identical heads have mutually exclusive bodies, as in distributional clauses (see \cite{Gutmann2011} for formal details). Lifting this restrictions would necessitate to capture interactions between different worlds as convolutions \cite{LucasPeterJ.F.2015MtIb}.

Contrary to Distributional Clauses, however, HAL-ProbLog allows for defining one and the same discrete random variable multiple times, which results in effectively encoding a noisy-or gate. On this stance, HAL-ProbLog inherits its semantics from (a)ProbLog. 

We are now able to interpret the HAL-ProbLog example program: a situation is modeled where the machine breaks down given that the temperature rises above 30 degrees or given that there is no cooling and the temperature rises above 20 degrees. The probability density modeling the temperature depends on whether it is hoy or not.

Let's move on by defining the semantics of the $\mathtt{valS}/ 2$ predicate. In the example in Eq.  \ref{exmp:halproblog} we saw that one and the same random variable can be distributed according to different distributions $f_i$, given mutually exclusive bodies $\mathtt{b}_i$. This entails that the value of the random variable depends on the world one is in. If we have now a predicate $\mathtt{valS}(\mathtt{t},T)$, then we accommodate for this fact by allowing the logic variable $\mathtt{T}$ to unify with all symbolic values for $\mathtt{t}$: $\mathtt{T} / \mathtt{t}_i|\mathtt{b}_i$. In other words: $\mathtt{T}$ unifies with the values of  any $\mathtt{t}_i$. The different $\mathtt{t}_i$'s are distinguishable by their mutually exclusive bodies $\mathtt{b}_i$. In our example, $T$ unifies with the value of the temperature given that it is a hot day and given that it is not a hot day.

Now we define the semantics of the $\mathtt{conS}/ 1$ predicate. Suppose that we have in a body of a clause an Iverson predicate $\mathtt{conS}(C)$, where the algebraic condition depends on a set of logic variables $\{ \mathtt{V}_i \}$. 
Then we add, for each ordered set of symbolic values $\{ \mathtt{v}_i \}$ that unifies with the ordered set of logic variables $\{ \mathtt{V}_i \}$ a clause of the following form to the program
\begin{align}\label{eqn:conS_ground}
\textstyle \left(C \theta_h, \mathcal{F}^S_{\mathtt{v}_i|\mathtt{b}_i} \right)::\mathtt{conS}(C \theta_{iv}) \leftarrow
\bigwedge\limits_{i} \mathtt{b}_i.
\end{align}
Where we have the substitutions $\theta_{h}= \{ \mathtt{V}_i / \mathtt{v}_i| \mathtt{b}_i \}$ and $\theta_{iv}= \{ \mathtt{V}_i / \mathtt{v}_i \}$. The $\mathtt{b}_i$'s are the bodies of clauses that have as head the random variables $\mathtt{v}_i$, respectively. $\mathcal{F}^S_{v_i|b_i}$ is short-hand for the set of distributions  $\left \{ \mathtt{v}_i|\mathtt{b}_i \sim f_i \right \}$.

The following example shows the effect of this transformation on the program in Eq. \ref{exmp:halproblog}.\footnote{The program in Eq. \ref{exmp:haproblog_con_ground} is also labeling atoms in the heads of clauses, which is supported by the aProbLog implementation and which can easily be eliminated. (just introduce a new predicate  $q(X)$ that returns $X=n$ if the $n$-th condition is true, and then replace the occurrences of $conS(Y)$ by $conS(Y,q(N))$).}
\begin{flalign}
&0.2\text{::}\mathtt{h}. \quad 0.01\text{::}\mathtt{no\_cool}.& \label{exmp:haproblog_con_ground} \\
&(\mathtt{t|\neg h}>20, \mathtt{normal}_\mathtt{t|\neg h}(20,5))\text{::}\mathtt{conS}(\mathtt{t}>20)\leftarrow \mathtt{\neg h}.& \nonumber \\
&(\mathtt{t|w}>20, \mathtt{normal}_\mathtt{t|w}(27,5))\text{::}\mathtt{conS}(\mathtt{t}>20)\leftarrow \mathtt{h}.& \nonumber \\
&(\mathtt{t|\neg h}>30, \mathtt{normal}_\mathtt{t|\neg h}(20,5))\text{::}\mathtt{conS}(\mathtt{t}>30)\leftarrow \mathtt{\neg h}.& \nonumber \\
&(\mathtt{t|w}>30, \mathtt{normal}_\mathtt{t|w}(27,5))\text{::}\mathtt{conS}(\mathtt{t}>30)\leftarrow \mathtt{h}.& \nonumber \\
&\mathtt{broken}\leftarrow \mathtt{conS}(\mathtt{t}>30). \; \mathtt{broken}\leftarrow \mathtt{no\_cool}, \mathtt{conS}(\mathtt{t}>20). & \nonumber 
\end{flalign}

\begin{figure}
\begin{center}
\includegraphics[width=\linewidth]{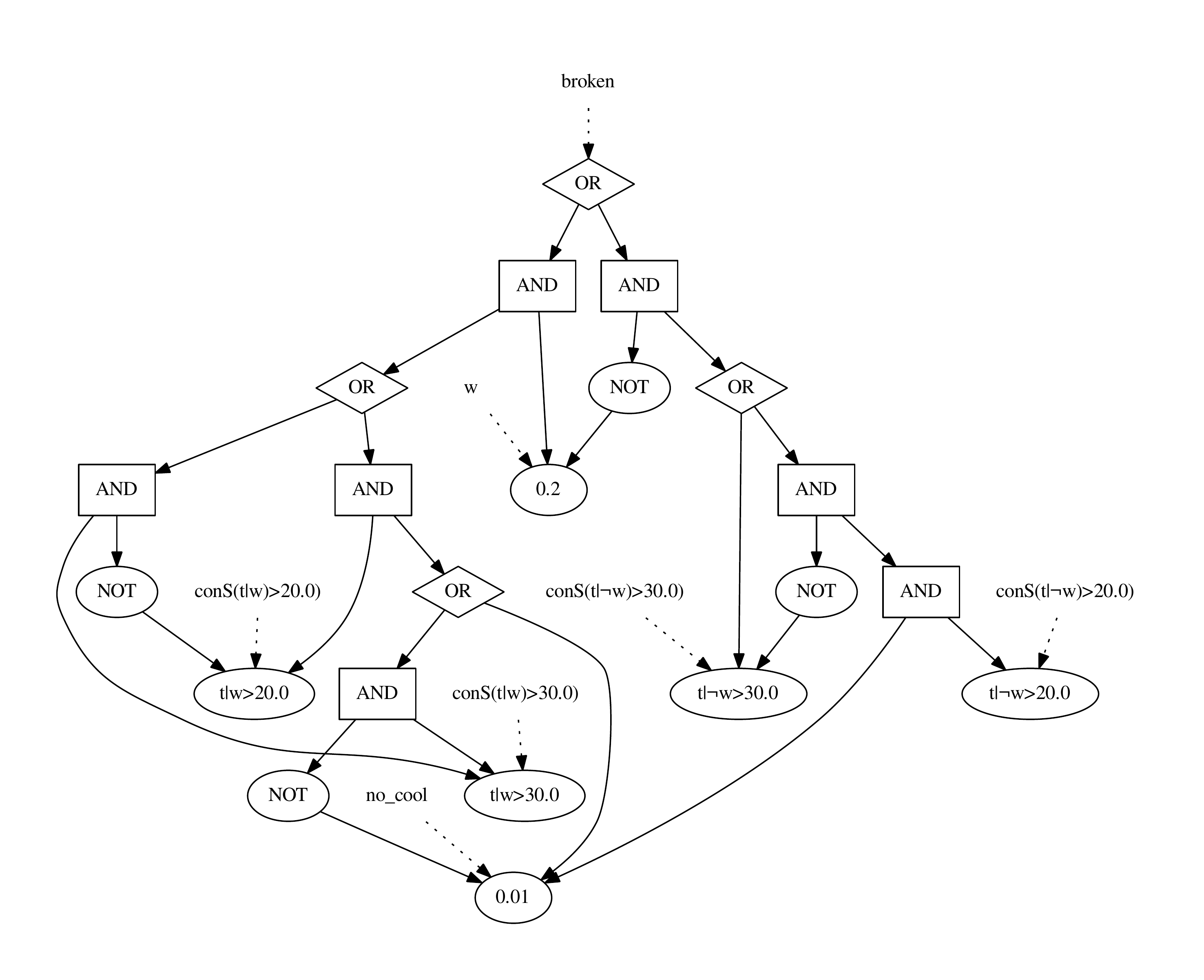}
\caption{Graphical representation of the HAL-ProbLog program in Eq. \ref{exmp:halproblog} compiled into an SDD.}
\label{fig:hal_problog_sdd_broken}
\end{center}
\end{figure}

We see that the Iverson predicate now functions as a literal. The transformation is finalized by removing the clauses whose head is a probability density from the program as they are no longer needed. This transformation is integrated within the aProbLog grounder (cf. section \ref{sec:implementation}). Compiling the program in Eq. \ref{exmp:haproblog_con_ground} into an SDD and calculating the probability of $\mathtt{broken}$ yields the following expression: 
\begin{align}
&(1-0.2)\left[ 0.01 \textstyle \int_{20 \leq 30} \mathcal{N}_\mathtt{t}(20,5) d\mathtt{t}+  \int_{30 >\mathtt{t}} \mathcal{N}_\mathtt{t}(20,5) d\mathtt{t}\right] \nonumber \\
& + 0.02\left[ 0.01 \textstyle \int_{20 <\mathtt{t}\leq 30} \mathcal{N}_\mathtt{t}(30,5) d\mathtt{t}+  \int_{30 >\mathtt{t}} \mathcal{N}_\mathtt{t}(30,5) d\mathtt{t}\right] \nonumber
\end{align}

Grounding the Iverson predicates in a HAL-ProbLog program actually results in an aProbLog program. Therefore, we say that a HAL-ProbLog program $P$ is valid if grounding all Iverson predicates results in a valid aProbLog program, given the probability density semiring $\mathcal{S}$.

\subsection{Implementation}\label{sec:implementation}

We implemented HAL-ProbLog as an extension of the publicly available ProbLog2 system~\cite{dries2015problog2}\footnote{\url{https://dtai.cs.kuleuven.be/problog/}}. This system provides a state-of-the-art implementation of the ProbLog language which is accessible as a module in Python.  The ProbLog system evaluates ProbLog models through the following pipeline: (1) reads the ProbLog model and transforms it into a clausal database representation, (2) generate a propositional and-or-graph representing the queries and evidence in the model using a built-in Prolog-based grounding engine, (3) transform the and-or-graph by removing cyclic dependencies (optional), (4) compile the model into an evaluatable form using knowledge compilation. Therefor we used SDDs, which are a subset of d-DNNF, as a target representation\footnote{\url{http://reasoning.cs.ucla.edu/sdd/}} 
\cite{choi2013dynamic}. (5) perform weighted model counting on this formula to obtain the final probability of interest.  
The ProbLog2 system also provides an implementation of aProbLog by allowing custom semirings to be defined in the final step of the process.

In order to implement our system we extended the grounder (ProbLog's step (2)) with support for the $\mathtt{conS/ 1}$ built-in that allows us to create dependencies of discrete random variables on continuous random variables.  It adds a $\mathtt{conS}$ node to the ground formula for all possible combination of ground literals that the algebraic condition involved depends on, cf. Eq. \ref{eqn:conS_ground}. Moreover, our system deploys Symbo in ProbLog's step (5) for evaluating compiled SMT(${\cal NRA}$) theories originating from hybrid HAL-ProbLog programs.

\section{Experimental Evaluation}
The question we would like to answer during the experimental evaluation is the following: How does solving hybrid probabilistic programs using Symbo, a logico-symbolic solver, compare to a pure, state-of-the-art, symbolic solver?

We answer this question by comparing HAL-ProbLog which uses Symbo to pure symbolic inference with the PSI-Solver in its native language. We compared Symbo and the PSI-Solver on the set of benchmark experiments given in section F of the Appendix in \cite{gehr2016psi}.\footnote{cf.: Fun \cite{InferNET14} and R2 \cite{nori2014r2}}

Experiments were performed on a laptop Intel(R) i7 CPU 2.60GHz with 16 Gb memory.

In Table \ref{table:symbo_psi}, we observe that Symbo outperforms the PSI-Solver on $9/10$ benchmarks, on $7/10$ even when including the time spend on the knowledge compilation step. Only for the ClickGraph benchmark PSI performs better than Symbo, which timed-out after $15$s during the integration step. This is because PSI integrates out variables after loop iterations. This is not yet supported in HAL-ProbLog and Symbo ends up with a large symbolic expression that is hard to integrate over. This could be solved by, for example, using sub-queries in HAL-ProbLog, as can be done in ProbLog2.

It is generally beneficial to perform logical inference on top of symbolic inference in the hybrid and thereby also in the discrete domain.
\begin{table}[ht]
\begin{center}
\resizebox{\columnwidth}{!}{
\begin{tabular}{l|ll|l|l }
   Benchmark            & KC & Evaluation   & PSI & Domain \\
\hline
BurglarAlarm  & $31.4$ & $0.8$ &   $190.1$ & D \\
CoinBias        & $41.9$ & $7.9$  &  $12.9$ & H \\
Grass            & $31.2$ & $1.2$  &   $228.0$ & D \\
NoisyOR      & $35.8$ & $11.2$ &   $12.7$ & D \\
TwoCoins      & $27.0$ & $2.1$ &    $57.8$ & D \\
ClickGraph      & $4300$ & -- &   $10500$  & H \\
ClinicalTrial      & $54.6$ & $25.7$   &  $3400$ & H \\
AddFun/max      & $25.2$ & $4.4$ &    $53.1$ & H \\
AddFun/sum      & $27.1$ & $2.1$ &   $84.9$ &  H \\
MurderMystery      & $27.6$ & $0.3$   &  $65.4$ & D \\
\end{tabular}
}
\caption{Knowledge compilation and arithmetic circuit evaluation times for Symbo, and problem solving time for PSI. Times are given in ms. Run times were averaged over 50 runs. The domain column indicates whether the problem is {\bf D}iscrete or {\bf H}ybrid.}
\label{table:symbo_psi}
\end{center}
\end{table}

\section{Related Work}
While knowledge compilation with SDDs and other representations has been used for (WMC) in probabilistic graphical models \cite{Choi2013} and probabilistic logic programming \cite{VLASSELAER201615}, it has to the best of our knowledge, not yet been applied to support hybrid exact inference.

W.r.t. hybrid inference in probabilistic programming, there are basically two classes of approaches: approximate and exact.
Firstly, for what concerns exact inference, there is the already mentioned work for imperative probabilistic programming \cite{gehr2016psi}, which has contributed the PSI solver that we use in Symbo. Furthermore, our work shows that knowledge compilation can speed up the inference in PSI and that the resulting framework applies hybrid probabilistic logics, too. 
Another approach related to exact inference in probabilistic logic programming is that of \cite{islam2012inference}.  Similarly to Symbo, they symbolically evaluate a theory in order to obtain an expression for a probability density. However, their approach is restricted to Gaussian densities and more importantly it is built on top  of Prism \cite{sato1995statistical}, which assumes that proofs are mutually exclusive, and which avoids the disjoint sum problem. As a consequence they do not support WMI in its full generality. Supporting WMI requires the KC step, which is not addressed in their work. 

Secondly, for what concerns approximate inference, we have the sampling approaches in distibutional clauses by   \cite{gutmann2011magic,nitti2016probabilistic} and BLOG \cite{MMRSOK07}. 
For   Distributional Clauses, one uses importance sampling to sample from probability distributions and densities alike, combined with likelihood weighting.

Approximate inference is also performed in \cite{michels2016approximate}. In their work, a hybrid probabilistic problem is represented by so called \textit{hybrid probability trees}. A node in the tree can then split up a continuous variable on an arbitrary value and for each child of the node an upper and lower probability bound can be calculated, which then gives upper and lower probability bounds at the splitting node. Going deeper in the tree yields tighter and tighter bounds. 

Finally, there is the work on inference in  weighted model integration \cite{belle2015probabilistic,morettin2017efficient}, which  handles probability densities by splitting them up (and thereby approximating) into piecewise polynomials and then carrying out exact inference.  This is somewhat related also to \cite{Gutmann2011}, who pursued a similar procedure by restricting distributions to Gaussians which can be chopped up into easily integrable pieces. In contrast to these works, we provide a much larger class of densities and constraints.

In this line of work \cite{belle2016component} also investigated component caching while performing a DPLL search when calculating a WMI and DPLL search is indeed related to knowledge compilation. However, the method proposed in their work is strictly limited to piecewise polynomials. We, again, completely lift this restrictions and are able to perform WMI via knowledge compilation on SMT($\mathcal{NRA}$) formulas using probability density functions instead of piecewise polynomials on SMT($\mathcal{LRA}$).

\section{Conclusion}
We have shown how knowledge compilation can be applied to the task of weighting model integration by leveraging algebraic model counting.   We have also introduced an effective logico-symbolic solver based on this idea. Finally, we presented HAL-ProbLog, a probabilistic logic programming language that is capable of  fully harnessing the logical structure underlying a hybrid probabilistic program through KC in the hybrid domain. 

In future work we would like explore non-factorized weight functions in the context of knowledge compilation and weighted model integration. Especially, as these non-factorized weight functions are presently predominantly used, cf. \cite{morettin2017efficient}.

\bibliography{references}
\bibliographystyle{aaai}

\end{document}